\pdfoutput=1
\documentclass[11pt]{article}

\usepackage[preprint]{acl}

\usepackage{times}
\usepackage{latexsym}
\usepackage{colortbl}
\usepackage{xcolor}
\usepackage{pgf}
\usepackage{subcaption}
\usepackage{float}
\usepackage{hhline}
\usepackage{booktabs}
\usepackage{multirow}

\newcommand{\heatmap}[1]{%
    \pgfmathsetmacro{\cellcolorvalue}{int(10 + #1*90/100)}
    \edef\temp{\noexpand\cellcolor{blue!\cellcolorvalue}}\temp #1
}

\usepackage[T1]{fontenc}
\usepackage[utf8]{inputenc}

\usepackage{microtype}

\usepackage{inconsolata}

\title{On  Evaluation Protocols for Data Augmentation\\ in a Limited Data Scenario}
\author{Frédéric Piedboeuf \\
  Université de Montréal, RALI \\
  \texttt{frederic.piedboeuf@umontreal.ca} \\\And
  Philippe Langlais \\
  Université de Montréal, RALI \\}

\begin{document}
\maketitle

\begin{abstract}
Textual data augmentation (DA) is a prolific field of study where novel techniques to create artificial data are regularly proposed, and that has demonstrated great efficiency on small data settings, at least for text classification tasks. In this paper, we challenge those results, showing that classical data augmentation (which modify sentences) is simply a way of performing better fine-tuning, and that spending more time doing so before applying data augmentation negates its effect. This is a significant contribution as it answers several questions that were left open in recent years, namely~: which DA technique performs best (all of them as long as they generate data close enough to the training set, as to not impair training) and why did DA show positive results (facilitates training of network). We further show that zero- and few-shot DA via conversational agents such as ChatGPT or LLama2  can increase performances, confirming that this form of data augmentation is preferable to classical methods. 
\end{abstract}

% =====================
% =====================

\section{Introduction}

\newcommand{\tabcolOneLen}{3cm}
\newcommand{\tableColLen}{5cm}
\newcommand{\vspaceneg}{\vspace{-10px}}
\newcommand{\sig}[1]{\underline{#1}}

% \section{Introduction}
Data augmentation (DA) consists in generating artificial data points with the hope of improving the training of a model. In this paper, we focus on interpretable textual DA (methods that generate new sentences) for text classification in a limited data scenario, a practical setting of interest \cite{chen_empirical_2021}. Research generally finds that DA provides a great increase on small data settings~\cite{chen_empirical_2021, kumar_data_2021}, a small increase in classification with medium datasets (up to 1000 examples)~\cite{karimi_aeda_2021, liesting_data_2021}, and almost no increase on large datasets~\cite{kobayashi_contextual_2018, yang_generative_2020,okimura-etal-2022-impact}.  

%DA algorithms proposed in the literature have an important effect not only because they guide practitioners towards one technique or the other, but also because they contribute to the understanding of how neural networks learn. Showing, for example, that a word substitution algorithm helps a pretrained model like BERT \cite{Devlin} has important implications about its inner working.  

In this paper, we show that existing experimental protocols in DA studies are misleading. In particular, inadequate fine-tuning of the baseline (model trained without DA) from previous studies lead to an overestimation of the impact of textual DA, and training the models for longer results in an absence of gains with previous DA methods. 
%In fact, we show that by simply playing with the patience parameter as well as occasionally adding label smoothing~\cite{szegedy_rethinking_2015}, we can raise the performance of the baseline classifier to be as good as performances with DA, and that when the classifier is trained correctly DA does not bring improvements. 
%
We also consider more realistic protocols for DA evaluation, where we either don't have access to clean validation data, or where we adjust the training/validation split to better reflect real-world conditions.
Lastly, we compare newer DA methods using ChatGPT~\cite{ouyang_training_2022} and Llama2~\cite{touvron_llama_2023} and show that the only thing that can consistently increase classification performance is generating data akin to external data (with zero or few-shot data generation), rather than generating new data from the current training distribution (by paraphrasing or modifying training sentences), as classical DA approaches do.

% Inspired by recent papers questioning the methodology of weakly supervised learning studies~\cite{zhu_weaker_2023}, we revisit DA experiments for classification and test the DA methods with an updated training methodology, focussing on the small data setting for which gains have been the largest. We show that by simply playing with the patience parameter of the classifier as well as occasionally adding label smoothing~\cite{szegedy_rethinking_2015}, we can raise the performance of the base classifier to be as good as performances with DA, thus negating the benefits from DA. We further integrate novel DA methods using ChatGPT and Llama2 and show that the only thing that can consistently increase classification performance is generating data akin to external data, rather than generating new data from the current training distribution, as classical DA approaches do. 

% \begin{enumerate}
%     \item We show that previous gains from interpretable textual data augmentation are due to a lack of fine-tuning of the network,
%     \item We demonstrate that ChatGPT and Llama2 can be used to generate external data which can increase the performance, but that this does not perform as well as collecting and annotating new data,
%     \item We test several methods of generating data with ChatGPT/Llama2 and show that using k-shot generation is more efficient than paraphrasing or zero-shot generation,
%     \item We show that \texttt{Llama2-13B-Chat}, while being free of use, is less efficient than ChatGPT for data augmentation.
% \end{enumerate}

Overall, this paper covers several important contributions. The first and most important of them lies in showing that classical data augmentation (DA methods from before the advent of very Large Language Models, or LLMs) does not work on textual classification, and we also answer why previous studies showed positive results. These positive results have been a big question in recent years, as nobody could explain with satisfaction why generated sentences helped transformers learn better, the best hypothesis being that it brought some kind of regularization to the network~\citep{feng_survey_2021, queiroz_abonizio_pre-trained_2020}.

Our second contribution is a questioning of the use of the validation data in data augmentation for small data settings. DA research generally supposes the availability of clean validation samples, an often unrealistic setting. With the splits used in those studies, one finds themself with a few tens of data points for training the model and a few hundreds for validation. In this paper, we  consider better splitting and more realistic uses of data, showing that it is often more advantageous to use all available data as training data and to fine-tune for a longer time.

The third contribution is showing that by using LLMs to generate novel sentences (and not just paraphrasing existing ones), the performance increases, but that this is still not as efficient as manually collecting and annotating novel data. We furthermore analyze two strategies for generating novel data, namely zero-shot data generation (using the task description) and 3-shot data generation (giving three examples to the LLM), showing that in most cases the 3-shot strategy largely outperforms zero-shot generation.

The last contribution joins a small but growing literature comparing LLMs in showing that ChatGPT outperforms Llama2~\citep{liu_radiology-llama2_2023, platek_three_2023, guo_is_2023}. To our knowledge, we are the first to study this in the context of DA.

The paper is organized as such. Section~\ref{Sec:RelWork} goes over the literature on interpretable textual DA. Sections~\ref{Sec:Datasets} and~\ref{Sec:DAStrat} explain respectively the datasets and the DA methods we compare. Section~\ref{Sec:Protocols} describes our evaluation protocols, then Section~\ref{Sec:Results} presents our results. We discuss our work in Section~\ref{Sec:Discussion}.

% =====================
% =====================

\section{Related Work}
\label{Sec:RelWork}

We may categorize DA techniques in three broad families~: word-level operations, paraphrasing, and generative DA. We note that this is not an absolute categorization, and each of these families regroup several families of techniques. We also refer to techniques that were developed before LLMs as \textit{classical DA}. These modify the starting sentences in some ways, bringing variations to the dataset. 

The first family does so by affecting individual words. The most seminal technique of this category is EDA, which considers four operations : word substitution, word deletion, word swapping, and insertion of related vocabulary~\citep{wei_eda_2019, liesting_data_2021}. For insertion and substitution, EDA uses WordNet~\cite{miller_wordnet_1998} to find synonyms of words from the input sentence. Another simple word-level technique is AEDA~\cite{karimi_aeda_2021}, where random punctuation is inserted in between words of the sentence. 

Other ways of replacing words have also been considered, such as the use of a pre-trained neural network to predict masked words~\cite{kobayashi_contextual_2018}, using pre-trained embeddings to find words close in the embedded space~\cite{marivate_improving_2020}, or using BERT/BART conditioned on the class to predict masked words or spans~\cite{wu_conditional_2019, kumar_data_2021}. While easy to implement, those techniques tend to bring little diversity.

The second family of techniques acts at the sentence level, by considering the whole sentence for creating paraphrases. The seminal technique representing this family is Back-translation (BT), in which a sentence is translated to another language and then back into English~\citep{hayashi_back-translation-style_2018, yu_qanet_2018, corbeil_bet_2020, edunov_understanding_2018}. The use of generative models for paraphrasing has also been considered, for example by encoding and decoding a sentence through a VAE~\citep{mesbah_training_2019, yerukola_data_2021, nishizaki_data_2017}.  Some other strategies that have been considered are modifying syntactic trees~\cite{coulombe_text_2018}, fine-tuning BART on an in-domain corpus of paraphrases~\cite{okur_data_2022}, using an off the shelf T5 model for paraphrasing (T5-Tapaco)~\cite{piedboeuf_is_2023}, or using LLMs such as ChatGPT for paraphrasing~\cite{fang_chatgpt_2023}.

Finally, generative methods aim to generate novel sentences from the same distribution as the training data. With an ideal generative strategy, this is equivalent to collecting new data, without the annotation cost associated to the collection process. Early studies of these techniques used GPT-2~\citep{kumar_data_2020, liu_data_2020, yang_generative_2020, bayer_data_2023} or VAEs~\cite{qiu_easyaug_2020, piedboeuf_effective_2022}, but recently focus has shifted to the use of LLMs, first with GPT3~\citep{yoo_gpt3mix_2021, sahu_data_2022}, and then with ChatGPT for zero or small data generation~\citep{moller_is_2023, ubani_zeroshotdataaug_2023, shushkevich_tudublin_2023, sharma_team_2023}.  

Some recent works pertinent to this paper are the research of~\cite{kim_alp_2022, zhu_weaker_2023}.~\citet{kim_alp_2022} question the training-validation splitting methodology in semi-supervised learning, showing that it is more efficient to fine-tune on augmented samples (created with DA) and use the original training sentences as validation data, instead of using a classical training/validation split. In~\cite{zhu_weaker_2023}, the authors note that weakly supervised learning approaches are evaluated by assuming the availability of clean validation samples, which is not often the case when working with small data. They notably develop novel methods for testing training under more realistic small data learning settings. In Section~\ref{Sec:Protocols}, we take inspiration of those papers to define our new protocols.

% =====================
% =====================

\section{Datasets}
\label{Sec:Datasets}

We use five popular datasets to test the various data augmentation strategies : SST-2, Irony, IronyB, TREC6, and SNIPS. SST-2~\citep{socher_recursive_2013} is a dataset of movie review classification with the binary classes of positive or negative. Irony and IronyB~\citep{van_hee_semeval-2018_2018} are the binary and multiclass version of an Irony detection dataset. In the binary task (Irony), one must detect if tweets are ironic or not, and in the multiclass version one must as well determine which type of Irony the tweet represents, if the tweet is ironic (between polarity clash, situational irony, and other irony). TREC6~\citep{li_learning_2002} is a task of question classification, where questions must be separated into six classes (abbreviation, description, entities, human beings, locations, and numeric values). Finally, SNIPS~\cite{coucke_snips_2018} is a dataset of intent classification, where short commands have to be classified into different intents such as PlayMusic or  GetWeather. Some characteristics of the datasets are available in Table~\ref{tab:datasets}.

\setlength{\tabcolsep}{2.8pt}
\begin{table}[h]
\renewcommand{\arraystretch}{1.1}
    \centering
    \begin{tabular}{lrrrrr} \hline
        Name  & SST2 & Irony &  IronyB & TREC6 & SNIPS\\ \midrule\midrule
        |classes| & 2 & 2 & 4 & 6 & 7\\ 
       sent. len. & 19.3 & 13.7   & 13.7& 10.2 & 9.3  \\  \midrule
        |train| & 6920 & 2683 & 2681 & 5452 & 13084 \\  
        |val| & 872 & 460 & 460 & 500 & 700\\  
        |test| & 1821 & 3834  & 3832 & 492 & 700 \\ \bottomrule
    \end{tabular}
    \caption{Characteristics of the classification tasks tackled in this study. The length of the sentences is defined by the number of white-space separated tokens.}
    \label{tab:datasets}
    \vspaceneg{}
\end{table}

Due to lack of computing budget\footnote{As we carefully fine-tune every baseline and run every experiment 15 times, this demands a significant time investment.} we leave the investigations of other tasks such as those considered in \cite{chen_empirical_2021} to future work.

% =====================
% =====================

\section{Data augmentation methods}
\label{Sec:DAStrat}

\begin{figure}
    \centering
    \includegraphics[scale=0.4]{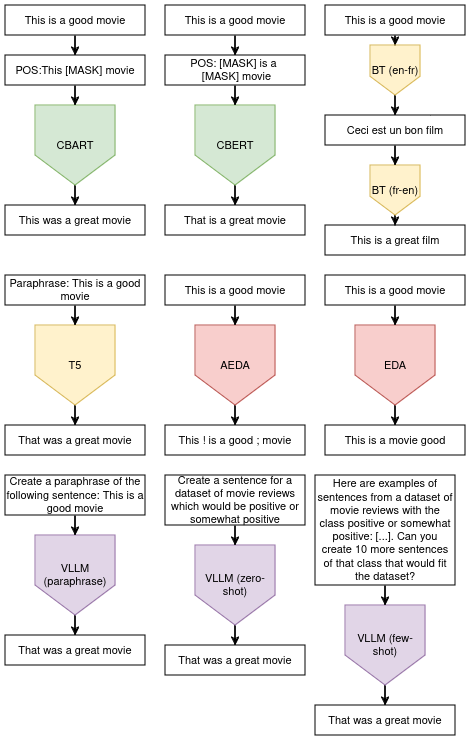}
    \caption{Strategies tested in this paper. Green algorithms are Contextual-based methods, yellow are paraphrasing methods, red are word-manipulation methods, and purple are methods using LLMs.}
    \label{fig:DeatDA}
    \vspaceneg{}
\end{figure}

While DA methods are continuously being proposed, objective evaluation is difficult due to the lack of extensive comparison studies. We point the reader to~\cite{chen_empirical_2021} for a literature review on DA methods, and~\cite{ding-etal-2024-data} for one specific to generative approaches. 
Here, we compare LLM based methods to previous (classical) approaches. All strategies are illustrated in Figure~\ref{fig:DeatDA}. Code and all hyperparameters are available in the additional material.

\subsection{Classical methods}

\citet{kumar_data_2021, piedboeuf_is_2023} compared several ``classical''  DA methods, from which we select three families of strategies that have been shown to be efficient. Concretely, we test word-manipulation methods (EDA, AEDA), conditional contextual based methods (CBERT, CBART), and paraphrase based methods  (T5, BT).

EDA (multiple operations on words) and AEDA (insertion of punctuation) can be implemented simply and are resource efficient. Following experiments and results from the literature, we affect 10\% of the words of a sentence in EDA, and use the formula given in~\citep{karimi_aeda_2021} to calculate the number of punctuation signs to insert for AEDA.\footnote{EDA works by randomly selecting one of four operations (insertion of related words, word swapping, word deletion, and word substitution) and applying it to a percentage of the words of the sentence. AEDA works by simply inserting random punctuation (among "?", ".", ";", ":", "!", and ",") into the sentence.}

CBERT and CBART are more involved methods that leverage the masked words prediction task to generate new words conditionally on the class. Concretely, we mask words, prepend the class to the sentence (with a separation token), and fine-tune the model to predict the masked words. Generation of sentences follow the same process. The difference between CBERT and CBART is that latter can predict \textit{spans} of words, allowing it more flexibility in the generated sentences. 

% CGPT follows a similar process where the class is prepended to the sentences and GPT-2 is fine-tuned on the training set augmented with the class. For generation, we can then simply input the class in the network and let it generate novel sentences which should be of the same distribution. 

BT and T5-Tapaco aim to produce paraphrases of the original sentences to bring diversity to the training set. In BT, paraphrases are generated by translating the sentence into a second language and then back into English, and we use WMT\footnote{\url{https://huggingface.co/facebook/wmt19-De-en}} with German as a pivot language, which has shown good performances in the past~\cite{edunov_understanding_2018}. T5-Tapaco makes use of \texttt{T5-small-Tapaco}\footnote{\url{https://huggingface.co/hetpandya/t5-small-tapaco}}, which is a T5 model fine-tuned on the TaPaCo paraphrase corpus~\citep{scherrer_tapaco_2020}, allowing us to directly generate paraphrases of the sentences. 

\subsection{Large Language Models}

LLMs are ubiquitous in NLP, and have rapidly gained traction in data augmentation. Here, we test three standard strategies~: paraphrasing sentences (P), zero-shot generation (ZS) by giving the LLMs a description of the task, and a 3-shot (3S) generation strategy where we give examples of the given class in addition to the description. Exact prompts used are provided in appendix~\ref{Appendix:Prompts}. We test two models, ChatGPT and Llama2 (using \texttt{LLama-2-13B-Chat}) with the same prompts. 

Note that we only use LLMs to generate new data with which we train a classifier, as with any other method we compare. Using LLMs directly to label examples may lead to better classification for some tasks, but we leave this for future investigations. In any case, training a classifier is a practical solution for many problems of interest,  as well as a  cheaper option to deploy.

\subsection{Baselines}

Finally, we implement three simple baselines. The first one consists in training the network without the use of DA (denoted ``Baseline''). The second one is an idealized strategy (denoted ```Perfect'') where we fetch additional unused data from the training set to act as generated data. This gives an idea of the results obtainable should one collect and annotate data instead of using DA. 

The last baseline is a strategy which we denote ``Copy'', and which artificially inflates the size of the training set by copying multiple times (according to the ratio parameter) the original data, without modification. If the only effect of DA is --- as we suspect ---  to help fine-tune the network better, and the modifications brought by DA are not helpful for that, then the Copy strategy should be just as efficient as all other classical DA methods.

% =====================
% =====================

\section{Experimental setups}
\label{Sec:Protocols}
In this section we describe the experimental setups for our two main experiments, that is to say the testing of DA methods with better fine-tuning, and the development of better experimental protocols for DA. Our focus in this paper is data augmentation for small data, for which we use starting size of 10 and 20, but we also report results for medium training sizes (500 and 1000) in Appendix~\ref{Appendix:SuppResults}, a setting commonly explored in the literature. When subsampling to create the training set, we make sure to choose an equal number of data points for each class, by sampling more data if needed (e.g. the actual dataset size for TREC6 and starting size of 10 is 12; two sentences per class). There is no consensus regarding the ratio of generated-to-genuine examples, but we found in both the literature and experiments that smaller datasets benefited more from larger ratios, while larger for larger datasets, we did not observe improvements with ratios larger than one. Thus, we use a ratio of 10 for the small data settings and of one for the larger dataset sizes.\footnote{Based on our main results, we can hypothesize that this is simply due to pretrained models needing more time to be fine-tuned correctly on smaller datasets than medium or large ones, and the larger ratio leads to a larger number of batches, meaning more training time.}

\subsection{Better fine-tuning}

To define our training protocol, we look at the literature to understand what is usually done.~\citet{piedboeuf_is_2023} use variable patience\footnote{The number of epochs during which the validation performance doesn't increase before we stop the training process, and we then select the best model from previous epochs.} for fine-tuning, depending on the dataset and dataset size, between 5 and 20 epochs, and~\citet{kumar_data_2021} use 100 epochs of warmup followed by 8 epochs of fine-tuning from which they select the best model.~\citet{wei_eda_2019} use early stopping with patience of 3 epochs on CNN and RNNs. 

In contrast, we use a patience parameter of 50, using the validation set to find when to stop, and we fine-tune the learning rate and use label smoothing~\cite{szegedy_rethinking_2015} (a mean of introducing noise to the labels for regularization) before applying data augmentation, using grid search.\footnote{While label smoothing was introduced in this paper as a mean of helping better learning of multiclass tasks, preliminary ablations experiments point that it raises the performance of both augmented and non-augmented dataset equally, and that the conclusions of this paper would be the same with only changing the number of epochs and hyper-parameter tuning.}  We report accuracy for binary tasks and the macro-F1 for multiclass tasks.  

%\subsection{On better significance testing} 
%The norm in data augmentation literature is to give the average and standard deviation over a fixed number of runs. However, to confirm whether the occasional improvements are due to randomness of the model or the effect of DA, we include in this paper statistical tests. For each starting size (10, 20) and dataset, we perform a two-tailed paired t-test, comparing the algorithms to each other\footnote{We choose to do this instead of a Hotelling’s T-square to take into account the fact the DA might work on some datasets but not others. For example, one would intuitively expect synonym replacement to work better on sentiment classification than on Irony detection.}. As is standard, we assess the two distributions to be different if the p-value is lower than 0.05.

\subsection{More realistic uses of data} 
In academic papers, DA is often evaluated with the assumption that we have access to clean validation data, often in larger quantity than what is available for the training set. This validation data is used to fine-tune both the DA algorithm and the classifier, resulting in unrealistic settings for practitioners, who may not have access to validation data or want to use their data in better ways. 

We aim to test more realistic settings with the validation data, inspired by~\citet{kim_alp_2022, zhu_weaker_2023}, who study the same problems but in different settings (semi-supervised learning and weakly supervised learning) and then discuss the significance of the results for textual DA. While~\citet{kim_alp_2022, zhu_weaker_2023} findings are relevant to ours, our interest in the results differ. 
In both papers, the authors attempt to find the best way to use the available data. In our case, we test several settings we think are more realistic, but or interest is to know if DA  useful when there is little or no validation data for fine-tuning the augmentation method and the classifier.

\begin{figure}[h]
    \centering
    \includegraphics[scale=0.4]{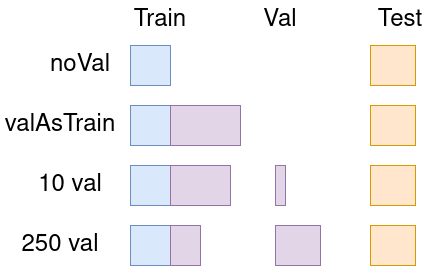}
    \caption{Graphic representation of the four settings we test for data augmentation on small data learning. Blue represent the original training set, purple the validation set, and yellow, the test set. }
    \label{fig:validation}
    \vspaceneg{}
\end{figure}

We redefine the train/val. split, using different strategies illustrated in Figure~\ref{fig:validation}. Firstly, we assume the validation data is not available (noVal), and we only have training data. In this setting, we train for a random number of epochs between 50 and 150, which is the range for which our models performed best in small data learning. This is inspired by the protocol of~\citet{zhu_weaker_2023}, where they randomly select a set of hyperparameters when no validation data is available.
Secondly, we assume the presence of validation data but redefine the train/val. split, by either 1- using it all as training data and training for a random number of epochs between 4 and 8 (valAsTrain), or 2- keeping some of it as validation data (10, 250) to fine-tune both DA algorithms and the classifier (10 val/250 val).

%We design experiments using two setups, electing to test the setups with the starting size of 10. This is motivated by the long time it takes to fine-tune and run all algorithms\footnote{Depending on the final training set size and the DA algorithms, running one experiment takes between 30 and 75 minutes, excluding the grid-search for the hyperparameters, on a RTX 3090.}, and the fact that the efficiency of DA was the same for both starting sizes, making duplication of the experiments unnecessary. An open question is how these settings would work if we had more initial training data, but we leave this for future work.
%

%Something worth noting is that all the protocols we know (including ours) use the validation data to fine-tune both the DA algorithm and the model to train. This is questionable, as keeping only 10 or 20 data points out of a few hundreds might not be the best way to utilize the data. While we keep this protocol in the main section to stay in accordance with the DA literature, we propose more realistic evaluation protocol of DA in Section~\ref{Sec:Ana}.

\newcommand{\lencol}{5cm}

% =====================
% =====================

\section{Experiments}
\label{Sec:Results}

In what follows, we add the augmented material to the (small) training set and fine-tune a BERT-Base model\footnote{We leave the study of other classifiers for future work.} in a supervised way for each classification task, a solution which has been demonstrated efficient~\cite{devlin_bert_2019}.

\subsection{On the need to better fine-tuning}

We first show the inefficiency of the fine-tuning protocol used in past DA studies by reporting the results they obtain without DA and comparing it to our results (without DA as well). We attempt to replicate the starting sizes, changing only the fine-tuning protocol to see the difference a longer patience, label smoothing, and better fine-tuning does. Table~\ref{tab:compfinetuning} presents the results of our experiments compared to those from the literature. Since the use of label smoothing depends on the dataset, we integrate its use as a hyperparameter.

%\addtocounter{table}{-2}
\begin{table}[h]
    \centering
    \begin{tabular}{l c c} \toprule
         & Reported & Ours  \\ \midrule
         \midrule
         Kumar+Ubani SST-2& 52.9 & 60.6 \\
         Kumar+Ubani SNIPS & 48.6 & 66.8 \\
         Kumar+Ubani TREC6 & 79.4 & 91.6\\
         Piedboeuf SST-2 & 87.7 & 87.7 \\
         Piedboeuf Irony & 65.6 & 67.0 \\
         Piedboeuf IronyB & 42.4 & 43.9 \\
         Piedboeuf TREC6 & 81.0 & 84.5\\ \bottomrule
    \end{tabular}
    \caption{Fine-tuning comparison from the literature and in this paper. All results are from training BERT without data augmentation, showing the difference made by using a longer patience as well as label smoothing.}
    \label{tab:compfinetuning}
    \vspaceneg{}
\end{table}

%\addtocounter{table}{1}
\setlength{\tabcolsep}{7pt}
\begin{table*}[]
    \centering
% \vspace{em}
    \begin{tabular}{p{2.5cm}rrrrrrr} \toprule
& SST2 & Irony & IronyB & Trec6 & SNIPS & $\,\,\,\,\,$& Average \\ \midrule \midrule
Baseline& 62.4/67.5& 54.1/61.9& 25.8/29.8& 34.4/43.3& 78.5/80.4& & 51.1/56.6\\
Perfect& \sig{82.3}/\sig{85.8} & \sig{62.4}/58.7& 19.3/33.7& \sig{72.5}/\sig{80.6} & \sig{94.0}/\sig{94.9}& & 66.1/70.7\\  \midrule
Copy& 61.0/68.1& 54.4/57.1& \textbf{25.5}/\textbf{30.3}& 36.4/\sig{49.3}& 77.7/\sig{83.4}& & 51.0/57.6\\ \\
EDA& 63.4/69.0& 54.9/\textbf{58.9}& 23.8/30.1& 35.8/\sig{48.0}& 76.6/\sig{83.2} & & 50.9/57.8\\
AEDA& 62.4/67.1& 55.4/56.9& 23.4/27.4& 33.2/45.7& 78.0/\sig{84.0} & & 50.5/56.2\\
BT& 60.2/66.8& 55.1/58.0& 24.7/27.9& 29.1/\sig{51.5} & 75.0/\sig{82.6}& & 48.8/57.4\\
CBERT& 62.5/65.4& \sig{56.3}/58.5& 23.2/28.3& 36.4/44.5& 76.3/\sig{83.4}& & 50.9/56.0\\
CBART& 62.5/65.6& 54.5/57.7& 23.6/29.0& 36.9/47.2& 77.0/\sig{84.5}& & 50.9/56.8\\
T5& 64.1/67.4& 56.3/58.3& 24.1/28.1& 34.3/\sig{48.5} & 76.7/\sig{84.1} & & 51.1/57.2\\ \\
GPT3.5-P& \sig{65.0}/68.3& \textbf{\sig{56.6}}/57.5& 22.7/28.3& 34.0/40.5& 80.2/\sig{84.7}& & 51.7/55.9\\
GPT3.5-ZS& \sig{82.0}/\textbf{\sig{78.7}}& \sig{56.4}/56.1& 23.5/24.6& \sig{40.0}/\sig{48.1} & \sig{84.8}/\sig{88.7}& & 57.3/59.2\\
GPT3.5-3S& \textbf{\sig{87.7}}/\sig{76.1} & 49.2/53.6& 20.9/24.6& \textbf{\sig{58.4}}/\textbf{\sig{63.9}}& \textbf{\sig{87.2}}/\textbf{\sig{89.6}}&& \textbf{60.7}/\textbf{61.5}\\
Llama2-P& \sig{66.3}/68.6& 55.3/57.1& 24.6/24.6& 33.4/41.6& 78.1/\sig{82.8}&& 51.6/55.0\\
Llama2-ZS& \sig{74.8}/\sig{74.0} & 54.3/57.2& 22.2/24.1& 37.3/47.3& 78.1/\sig{84.1}&& 53.4/57.3\\
Llama2-3S& 63.3/64.4& 54.1/54.8& 21.0/27.2& \sig{48.6}/\sig{55.8} & \sig{83.3}/\sig{88.2}&& 54.0/58.1\\
\bottomrule
    \end{tabular}
    \caption{Average metric over 15 runs for the training set sizes of 10  (left) and 20 (right) with a ratio of 1. We report accuracy for binary tasks and macro-f1 for multiclass ones. STDs are between 0.6 and 3.0, depending on the dataset. Results for which the difference with the baseline was found to be statistically significant according to a t-test are underlined.}
    \label{tab:resultsSmol}
    \vspaceneg{}
\end{table*}

As we can see, carefully fine-tuning the classifier increases its performance significantly, gaining between 2 and 10\%. Those gains are actually larger than the ones reported by authors while deploying DA. This aligns well with  our hypothesis that the gains observed in previous studies stem only from giving more time for the model to learn. Further, if the DA strategies involve simple transformations of the sentences, it is likely that BERT is not learning new information from these transformations, and that the same result will be achieved with the Copy strategy, which we analyze in the next section.

\subsection{On the inefficiency of classical DA}

We now turn to DA algorithms, showing that better fine-tuning leads to essentially useless (classical) DA algorithms. Indeed we argue that alleged gains from DA that were shown in the literature can be explained by inadequate training of the (initial) classifier. Results are presented in Table~\ref{tab:resultsSmol}. 

% The baseline correspond to results without data augmentation. 
% ALP : https://ojs.aaai.org/index.php/AAAI/article/view/21336

For the small data setting, we notice that, on average, only two strategies really outperform the baseline~: zero-shot and 3-shot generation (both using LLMs), 3-shot generation outperforming zero-shot, but not on all datasets. We explore this phenomenon in Section~\ref{Sec:AnaVLLMS}. The results for medium sizes (500 and 1000) are presented in Appendix~\ref{Appendix:SuppResults}, and are concordant with more recent literature showing a lack of improvement on such a setting. 

An important thing to note is that while some algorithms perform statistically better than the baseline at times, so does the ``Copy'' strategy. As the duplication of data should bring no new information, this suggests that the training protocol in previous studies was suboptimal and that the gains observed are not due to data augmentation. 

One avenue that could be explored to bridge the gap would be to use sampling with replacement while fine-tuning BERT, which should be equivalent to duplication of data.\footnote{There would still be minor differences as sampling with replacement would not represent each sample equally, but this difference should reasonably not have a large impact.} 
%\subsection{On Significance Testing}

Figure~\ref{tab:fewShotStats} provides the results of the statistical tests we conducted. To simplify reading, we group together all tests between the same algorithms, meaning all tests between algorithm A and algorithm B are compiled together. Each entry in the heatmap is therefore composed of 10 t-tests (5 datasets * 2 starting sizes), and we report the percentage of t-tests for which distributions were found to be statistically different, counting only the entries for which the row algorithm performs better than the column algorithm. This gives, for each row, the measure of how many times it beats the algorithm in the column by a statistically significant margin.\footnote{We provide the code along with the full saved results for all experiments, as well as the code necessary to run all statistical tests. See Figure~\ref{tab:fewShotStats500} in Appendix for medium data settings (500/1000).}

\setlength{\tabcolsep}{1pt}

\begin{figure}[]
    \centering
    \tiny
    \begin{subfigure}{0.48\textwidth}
        \centering
    \begin{tabular}{l*{16}{p{0.35cm}}}
         & \rotatebox{85}{Baseline} & \rotatebox{85}{Perfect} & \rotatebox{85}{Copy} & \rotatebox{85}{EDA} & \rotatebox{85}{AEDA}& \rotatebox{85}{BT} & \rotatebox{85}{CBERT}& \rotatebox{85}{CBART} & \rotatebox{85}{T5} & \rotatebox{85}{GPT3.5-P} & \rotatebox{85}{GPT3.5-ZS} & \rotatebox{85}{GPT3.5-3S} & \rotatebox{85}{Llama2-P} & \rotatebox{85}{Llama2-ZS} & \rotatebox{85}{Llama2-3S}\\
Baseline & \heatmap{0} & \heatmap{20} & \heatmap{10} & \heatmap{10} & \heatmap{10} & \heatmap{30} & \heatmap{10} & \heatmap{10} & \heatmap{10} & \heatmap{20} & \heatmap{20} & \heatmap{40} & \heatmap{10} & \heatmap{30} & \heatmap{30} \\
Perfect & \heatmap{70} & \heatmap{0} & \heatmap{70} & \heatmap{70} & \heatmap{70} & \heatmap{70} & \heatmap{70} & \heatmap{70} & \heatmap{70} & \heatmap{70} & \heatmap{70} & \heatmap{80} & \heatmap{80} & \heatmap{80} & \heatmap{80} \\
Copy & \heatmap{20} & \heatmap{10} & \heatmap{0} & \heatmap{0} & \heatmap{10} & \heatmap{20} & \heatmap{10} & \heatmap{0} & \heatmap{0} & \heatmap{30} & \heatmap{10} & \heatmap{40} & \heatmap{20} & \heatmap{20} & \heatmap{30} \\
EDA & \heatmap{20} & \heatmap{10} & \heatmap{20} & \heatmap{0} & \heatmap{10} & \heatmap{20} & \heatmap{0} & \heatmap{0} & \heatmap{0} & \heatmap{10} & \heatmap{20} & \heatmap{40} & \heatmap{20} & \heatmap{10} & \heatmap{20} \\
AEDA & \heatmap{10} & \heatmap{10} & \heatmap{0} & \heatmap{0} & \heatmap{0} & \heatmap{20} & \heatmap{0} & \heatmap{0} & \heatmap{0} & \heatmap{10} & \heatmap{0} & \heatmap{30} & \heatmap{20} & \heatmap{0} & \heatmap{0} \\
BT & \heatmap{20} & \heatmap{10} & \heatmap{0} & \heatmap{0} & \heatmap{10} & \heatmap{0} & \heatmap{10} & \heatmap{10} & \heatmap{0} & \heatmap{10} & \heatmap{10} & \heatmap{30} & \heatmap{10} & \heatmap{10} & \heatmap{20} \\
CBERT & \heatmap{20} & \heatmap{10} & \heatmap{10} & \heatmap{0} & \heatmap{10} & \heatmap{10} & \heatmap{0} & \heatmap{10} & \heatmap{0} & \heatmap{10} & \heatmap{0} & \heatmap{20} & \heatmap{10} & \heatmap{0} & \heatmap{20} \\
CBART & \heatmap{10} & \heatmap{10} & \heatmap{0} & \heatmap{0} & \heatmap{0} & \heatmap{10} & \heatmap{0} & \heatmap{0} & \heatmap{0} & \heatmap{10} & \heatmap{0} & \heatmap{30} & \heatmap{10} & \heatmap{10} & \heatmap{10} \\
T5 & \heatmap{20} & \heatmap{10} & \heatmap{0} & \heatmap{0} & \heatmap{0} & \heatmap{20} & \heatmap{10} & \heatmap{0} & \heatmap{0} & \heatmap{10} & \heatmap{0} & \heatmap{40} & \heatmap{10} & \heatmap{10} & \heatmap{20} \\
GPT3.5-P & \heatmap{30} & \heatmap{10} & \heatmap{10} & \heatmap{20} & \heatmap{10} & \heatmap{40} & \heatmap{20} & \heatmap{20} & \heatmap{10} & \heatmap{0} & \heatmap{10} & \heatmap{30} & \heatmap{10} & \heatmap{20} & \heatmap{20} \\
GPT3.5-ZS & \heatmap{70} & \heatmap{10} & \heatmap{60} & \heatmap{50} & \heatmap{50} & \heatmap{50} & \heatmap{60} & \heatmap{50} & \heatmap{50} & \heatmap{60} & \heatmap{0} & \heatmap{40} & \heatmap{60} & \heatmap{50} & \heatmap{30} \\
GPT3.5-3S & \heatmap{60} & \heatmap{10} & \heatmap{60} & \heatmap{60} & \heatmap{60} & \heatmap{60} & \heatmap{60} & \heatmap{60} & \heatmap{60} & \heatmap{60} & \heatmap{40} & \heatmap{0} & \heatmap{60} & \heatmap{60} & \heatmap{50} \\
Llama2-P & \heatmap{20} & \heatmap{10} & \heatmap{10} & \heatmap{0} & \heatmap{10} & \heatmap{20} & \heatmap{20} & \heatmap{10} & \heatmap{0} & \heatmap{0} & \heatmap{0} & \heatmap{30} & \heatmap{0} & \heatmap{0} & \heatmap{10} \\
Llama2-ZS & \heatmap{30} & \heatmap{10} & \heatmap{20} & \heatmap{20} & \heatmap{20} & \heatmap{30} & \heatmap{20} & \heatmap{20} & \heatmap{20} & \heatmap{30} & \heatmap{0} & \heatmap{20} & \heatmap{40} & \heatmap{0} & \heatmap{30} \\
Llama2-3S & \heatmap{40} & \heatmap{0} & \heatmap{40} & \heatmap{40} & \heatmap{40} & \heatmap{40} & \heatmap{40} & \heatmap{40} & \heatmap{40} & \heatmap{40} & \heatmap{20} & \heatmap{20} & \heatmap{40} & \heatmap{40} & \heatmap{0} \\
    \end{tabular}
    \end{subfigure}
    \caption{Percent of times the row algorithm performs statistically better than the column algorithm, with a p-value threshold of 0.05 and using a two-tails paired t-test, and across the two small data settings (10/20).}
    \label{tab:fewShotStats}
    \vspaceneg
\end{figure}

From the statistical tests, we note that only three methods outperform the others by a significant and consistent margin~: the perfect strategy, in which we simulate collecting and annotating external data, and the zero/k-shot generation with LLMs. All others perform equivalently and do not beat the baseline most of the time, or at least not by a significant margin. This suggests that the slight random variations we observe, which could be taken as a sign that some algorithms perform better than others, are due to randomness of the network and not a superior performance.

% =======================
\subsection{On more realistic uses of data}

%The preceeding section evaluated data augmentation algorithms using a standard training protocol. While our main result is that the efficiency of classical DA is generally over-estimated, this does not imply it is \textit{never} useful. 

%We design experiments using two setups, electing to test the setups with the starting size of 10. This is motivated by the long time it takes to fine-tune and run all algorithms\footnote{Depending on the final training set size and the DA algorithms, running one experiment takes between 30 and 75 minutes, excluding the grid-search for the hyperparameters, on a RTX 3090.}, and the fact that the efficiency of DA was the same for both starting sizes, making duplication of the experiments unnecessary. An open question is how these settings would work if we had more initial training data, but we leave this for future work.

As mentioned, the second point of interest is to evaluate DA in a more realistic setting with regard to the use of validation data. We select the following algorithms~: Copy, Perfect, EDA, AEDA, BT, CBERT, BART, T5, GPT3.5-ZS, and GPT3.5-3S. We use this limited selection for efficiency reasons, but also because our goal here is to establish whether DA helps at all, rather than knowing which DA technique works best. Finally, to keep in line with other DA studies, we use a ratio of 10 when the final training size is small (noVal) and of 1 for the other settings. Results are presented in table~\ref{tab:DAANalysisBettreVal}, and several important observations can be made. % unclear this selection thing

% \subsection{Use of data augmentation for allolanguages}
% A reminding question is whether data augmentation is efficient for non-english language, which we deem \textit{allolanguage} in this section. Multilingual systems often under-performed when compared to their english counterpart, and therefore there is still the chance that data augmentation does help the network. In fact, data augmentation is regularly used in classification in allolanguages (CITES).

% We select and adapt a few techniques that are easily transferable, namely AEDA, EDA, BT, and CBERT. We furthermore test the use of a VAE following~\cite{qiu_easyaug_2020, piedboeuf_effective_2022}, training one VAE per class. For testing our system, we select four datasets. We use SB10k, a dataset of sentiment classification from twitter in German~\cite{cieliebak_twitter_2017} (with the classes of negative, neutral, or positive tweets), SwaNews, a dataset of Swahili news classification\footnote{\url{https://zenodo.org/record/5514203\#.Y20HvblyZhE}} (local, international, finance, health, sports, and entertainment), koHateSpeech, a dataset of hate speech detection in Korean~\cite{moon_beep_2020} (hate speech, neutral, or offensive), and CLS, a document of French product review from Amazon~\cite{prettenhofer_cross-language_2010} (negative or positive). For the latter, since the reviews are much longer than the other tasks (average length of 103.8 words), we artificially shorten the entries by adding one sentence at the time until the length is more than 20 (tokenized at white space).

\setlength{\tabcolsep}{3pt}
\begin{table}[h]
    \centering
    \begin{tabular}{lcccc} \toprule
     & noVal & valAsTrain & 10 val & 250 val \\ \midrule \midrule
Baseline & 47.9 & 75.8 & 64.6 & 64.7 \\
Perfect & 68.8 & 82.4 & 69.4 & 69.9 \\  \midrule
Copy & 49.6 & \textbf{78.6} & \textbf{66.3} & 65.8 \\ \\
EDA & 49.1 & 77.8 & 66.0 & 64.9 \\
AEDA & 49.4 & 77.6 & 66.0 & \textbf{66.3} \\
BT & 44.8 & 75.9 & 64.6 & 65.0 \\
CBERT & 49.7 & 76.8 & 65.3 & 63.9 \\
BART & 50.6 & 76.6 & 65.3 & 63.9 \\
T5Par & 49.4 & 77.2 & 65.3 & 64.6 \\ \\ % never introduced
GPT3.5-ZS & 55.9 & 76.9 & 65.6 & 64.4 \\ 
GPT3.5-3S & \textbf{57.2} & 71.6 & 67.2 & 64.8 \\
\bottomrule
    \end{tabular}
    \caption{Average metrics over the five datasets for different settings and DA strategies. The best result for each setting is in bold.}
    \label{tab:DAANalysisBettreVal}
    \vspaceneg{}
\end{table}

First, in almost all cases and contrary to the main results of this paper, data augmentation reveals itself to be useful. However, the fact that the ``Copy'' strategy often outperforms the other strategies reinforces our belief that the use of data augmentation is simply to facilitate fine-tuning the network.

Second, there is no difference in performance between having 10 and 250 data points as validation data, with the rest as training set. This could be related to diminished increase in model performance as dataset size augments, or point to a trade-off in validation/training size and that a validation set of 10 data points is too small to be of use. 

Third, we confirm here that methods adding something akin to external data (GPT3.5-ZS and GPT3.5-3S) are the most useful techniques, but only on small training sizes. On larger ones, those methods do not bring any improvement whatsoever, most likely due to the general inefficiency of data augmentation in this setting.\footnote{This is concordant with the literature and our own experiments.} We also note that while the performance of GPT3.5-ZS is higher than GPT3.5-3S for two settings, this is due exclusively to its poor performance on the Irony datasets, which we discuss in Section~\ref{Sec:AnaVLLMS}. For the other datasets, GPT3.5-3S outperforms GPT3.5-ZS by a large margin.

Finally, it seems that the best strategy overall when the total amount of available data is only a few hundred is to use all of them as training data, with the Copy strategy.\footnote{Or, we can assume, a larger number of epoch.} This outperforms by a large margin the strategies where we keep some data as validation data, most likely because these strategies overfit. As mentioned, however, there is a large gap between 10 and 250 data and it is possible that some amount of data in the middle would perform better. We leave this investigation for future work. 

\subsection{Analysis of DA with LLMs}
\label{Sec:AnaVLLMS}

Our main results show that in a small data setting, 3-shot generation was often more efficient than zero-shot generation, by as much as 18\% in some cases. However, this does not hold for the two irony datasets, for which zero-shot outperforms 3-shot. In this section, we analyze the output of the algorithms, in the hope of understanding better why some prompt mechanisms work well on some datasets but not on others. 

We first show examples of generated sentences in Table~\ref{tab:GenSentIrony} and in Appendix~\ref{sec:gensentences}. For the sake of space, we only show generated sentences for the tasks of SST-2 and Irony, but we give all sentences generated by ChatGPT in the associated Github. 

 \parbox[t]{2mm}{} 
 
\begin{table}[h]
    \centering
    \begin{tabular}{ll l p{5.7cm}} \toprule
     & & \multicolumn{2}{l}{Couldn't have asked for a better night.} \\ \midrule
     & & -P   & Could not have hoped for a more perfect evening. \\ \smallskip  
      \multirow{5}{*}{\rotatebox[origin=c]{90}{Non-Ironic}} & 
         & -ZS & Wow, I absolutely love it when my phone battery dies in the middle of an important call. Just what I needed! \\ \smallskip 
     & & -3S & Enjoying a relaxing day at the beach with clear skies and warm sunshine. \#PerfectDay \\
     \toprule 
      & & \multicolumn{2}{l}{Work should be fun today :unamused\_face:} \\ \midrule
     & & -P   & Today at work is expected to be unenjoyable :unamused\_face: \\  \smallskip  
      \multirow{5}{*}{\rotatebox[origin=c]{90}{Ironic}}& 
        & -ZS & Oh, how wonderful! Another day of unexpected surprises and setbacks. It's like the universe's way of keeping me on my toes! \\ \smallskip  
     & & -3S & Declared I'd be more organized and lost my keys for the umpteenth time. I'm the queen of tidiness! :key::mag: \#OrganizationQueen \\
%     \bottomrule
    \end{tabular}
    \caption{Generated sentences by the GPT3.5-based DA methods for the Irony dataset for one ironic and one non-ironic example.}.
    \label{tab:GenSentIrony}
    \vspaceneg{}
\end{table}

By observing generated sentences, we see that ChatGPT has difficulty understanding the concept of Irony, as well as sticking to the more familiar language of Twitter. As such, adding more sentences doesn't help because the new sentences are too far from the training distribution to bring valuable information. Nevertheless, the 3-shot strategy does seem to bring the generated sentences closer to the training distribution, as we can at least observe that the new sentences contain hashtags and emojis. It is likely that by further fine-tuning the prompts, we would reduce the gap between the data distribution and the generated sentences and generate highly informative sentences.

\section{Conclusion}
\label{Sec:Discussion}

In this paper, we test DA for sentence classification and show that in both medium and small data learning, performances of DA had been overestimated by the use of inadequate training protocols.  Furthermore, while we have shown that  classical DA methods  are inefficient, we have also demonstrated that simulating external data collection with LLMs does improve the performance, and future work should therefore focus on this rather than techniques modifying genuine sentences. 

We want to emphasize that our results are only looking at \textit{balanced classification tasks}, and furthermore only at short text classification using BERT as a classifier. While this seems a small field of study, it is one of the most popular in textual DA literature, making our findings significant.

Future work should also evaluate whether the protocols are adapted for fields close to interpretable DA. Notably non-interpretable DA, or DA for unbalanced data, might still be an efficient mean of increasing performances. Other textual tasks might also still benefit from DA, due to different goals. As an example, data augmentation for the task of keyphrase generation~\citep{ray_chowdhury_kpdrop_2022, garg_data_2023} aims to encourage the network to generate more keyphrases that are absent from the input. It is very plausible that data augmentation may be useful here.

% As another example, data augmentation is sometime used to improve explainability~\citep{chen_improving_2020, ansari_data_2023}, and protocol should therefore be checked thoroughly in independent fields to ensure that it works as intended.

% In this paper, we look at the protocol surrounding data augmentation, and demonstrate that data augmentation is mainly used for easier fine-tuning, an effect that can be replicated by simply giving more care to the fine-tuning of the network, or by using a simple strategy where we duplicate the data.  We also consider other, more realistic applications of DA, where validation data is either not available or used as training data, instead of keeping hundreds of examples for validation and only a dozen for training the model.

Our findings are important as they answer many questions in textual DA which have been standing for years, namely which data augmentation algorithms is best (they all perform similarly except for data generation), why does it help (gives more time to the network to learn), and what makes a generated sentence informative (they don't bring contradictory information to the network). As noted, data generation remains one of the best way to perform data augmentation, if one can get the LLMs to generate data adequate for the training distribution.

\section{Limitations}
\label{Sec:Limitations}
As already stated, this study has a specific, although important, scope. We didn't look at the impact of data augmentation on other languages than English, or other textual tasks, and while we performed fine-tuning to the best of our knowledge, it is possible that some combinations of hyperparameters that were not explored may change some of our conclusions. We also omitted an ablation study of the fine-tuning (patience, label smoothing, grid search) due to time limitation, but this would provide more insight into the mechanism of training BERT. Furthermore, our exploration of hyperparameters for ChatGPT (mainly the temperature) was limited, and we only used the Llama2-13b-chat model as a comparison, but a comparison of other sizes and also other LLMs should be included in future work, such as Llama3~\cite{meta_meta_2024}, Gemini~\cite{gemini_team_gemini_2024}, GPT-4~\cite{openai_gpt-4_2024}, and others. Finally, we only looked at BERT as a classifier, which is the most common one used in DA studies, but it would be pertinent to not only look at other pretrained transformers but also at the impact of DA on more classical models such as CNNs, RNNs, or even statistical methods (SVM, NB, etc).

On a broader scope, there is no guarantee our results hold for other textual tasks (such as question answering, explainability, or keyphrase generation), which have different structures. A final point is that we only compared ourselves to the studies of~\citet{kumar_data_2021, piedboeuf_is_2023} to demonstrate the inefficiency of fine-tuning in past studies. While these are some of the most complete we found, it would be pertinent to repeat the experience with other papers.

\bibliography{ref}

\appendix

\section{Generated sentences}
\label{sec:gensentences}

\begin{table}[]
    \centering
    \begin{tabular}{c c c} \toprule
    \parbox[t]{2mm}{\multirow{12}{*}{\rotatebox[origin=c]{90}{Negative}}} & \multirow{3}{*}{Dataset} & \multirow{3}{*}{\parbox{\lencol}{apparently reassembled from the cutting-room floor of any given daytime soap.}} \\
         % & \\
         & \\
         & \\ \hhline{~==}
     & \multirow{3}{*}{GPT3.5-P} & \multirow{3}{*}{\parbox{\lencol}{It appears to have been pieced together from discarded scenes of any typical daytime soap opera.}} \\
         & \\
         % & \\
         & \\ \cline{2-3}
         &\multirow{3}{*}{GPT3.5-ZS} & \multirow{3}{*}{\parbox{\lencol}{The film was a major disappointment, lacking any coherent plot or engaging characters.}}\\
         % & \\
         & \\
         & \\ \cline{2-3}
         &\multirow{3}{*}{GPT3.5-3S} & \multirow{3}{*}{\parbox{\lencol}{It's a shame that such a promising concept was wasted on a poorly executed film.}} \\
         % & \\
         & \\
         & \\ \hhline{===}
         \parbox[t]{2mm}{\multirow{13}{*}{\rotatebox[origin=c]{90}{Positive}}}&\multirow{2}{*}{Dataset} & \multirow{2}{*}{\parbox{\lencol}{béart and berling are both superb , while huppert ... is magnificent.}} \\
         % & \\
         & \\ \hhline{~==}
         & \multirow{4}{*}{GPT3.5 - P} & \multirow{4}{*}{\parbox{\lencol}{Both Béart and Berling deliver exceptional performances, while Huppert... she is truly magnificent.}} \\
         & \\
         % & \\
         & \\
         
         & \\ \cline{2-3}
         &\multirow{3}{*}{GPT3.5-ZS} & \multirow{3}{*}{\parbox{\lencol}{This film is a refreshing breath of fresh air, offering a unique and original take on familiar themes.}} \\
         & \\
         % & \\
         & \\ \cline{2-3}
          &\multirow{4}{*}{GPT3.5-3S} & \multirow{4}{*}{\parbox{\lencol}{This film is a delightful and heartwarming celebration of friendship, full of laughter and touching moments.}} \\
         & \\
         % & \\
         
         & \\
         & \\ \midrule
% &GPT3.5-P  \\
%          &GPT3.5 -ZS \\
%          &GPT3.5 - 3S \\
%         &GPT3.5-P  \\
%          &GPT3.5 -ZS \\
%          &GPT3.5 - 3S \\
    \end{tabular}
    \caption{Examples of generated sentences for the SST-2 dataset.}
    \label{tab:GenSentSST2}
\end{table}
\section{Prompts}
\label{Appendix:Prompts}
In this section, we describe the prompts we used for ChatGPT and LLama2. For the paraphrasing strategy, we copy~\citet{piedboeuf_is_2023}, who simply asks for paraphrases in batch if the ratio is 1 and for multiple paraphrases if the ratio is larger than one. We refer to their paper for more detail on the prompting.

For the zero and three-shot generation, we use the template shown in Table~\ref{tab:PromptsGen}.

\begin{table}[]
    \hspace{-0.5cm}
    \centering
    \setlength{\tabcolsep}{2pt}
    \begin{tabular}{ll} \toprule
    \multirow{1}{*}{DATASET\_DESC} & \multirow{1}{*}{CLASS\_DESC}  \\ \midrule
        \multirow{2}{*}{movie reviews} & \cellcolor{gray!10}negative or somewhat negative  \\
         & positive or somewhat positive \\ \midrule
         \multirow{2}{*}{\parbox{\tabcolOneLen}{headline Fake/Real news classification}} & \cellcolor{gray!10}Real\\
          & Fake \\ \midrule
        \multirow{2}{*}{\parbox{\tabcolOneLen}{Ironic tweet detection}} & \cellcolor{gray!10} Non-Ironic Tweets \\
        & Ironic Tweets \\ \midrule
        \multirow{12}{*}{\parbox{\tabcolOneLen}{Ironic tweet detection}} & \cellcolor{gray!10}\multirow{1}{*}{\parbox{\tableColLen}{Tweets that are not ironic}}\\
        & \\
         & \\
         & \\
         &\multirow{-4}{*}{\parbox{\tableColLen}{Tweets ironic by polarity contrast, where the polarity is inverted between the literal and intended evaluation}}\\
                &\cellcolor{gray!10} \\
         &\cellcolor{gray!10} \\
         &\cellcolor{gray!10}\multirow{-3}{*}{\parbox{\tableColLen}{Tweets ironic by Situational Irony, where a situation fails to meet some expectation}}\\
                & \\
         & \\
         & \\
         &\multirow{-4}{*}{\parbox{\tableColLen}{Tweets ironic by Other type of Irony, where the Irony is neither by Polarity Contrast or by Situational Irony}}\\ \midrule
          \multirow{13}{*}{\parbox{\tabcolOneLen}{Question classification}} & \cellcolor{gray!10}\multirow{1}{*}{\parbox{\tableColLen}{Questions about an abbreviation}}\\
        & \\
         & \multirow{-2}{*}{\parbox{\tableColLen}{Questions about an entity (event, animal, language, etc)}}\\
         & \cellcolor{gray!10}\\
         &  \cellcolor{gray!10}\\
         &\cellcolor{gray!10}\multirow{-3}{*}{\parbox{\tableColLen}{Question concerning a description (of something, a definition, a reason, etc)}}\\
                   & \\
         &  \\
         &\multirow{-3}{*}{\parbox{\tableColLen}{Questions about a human  (description of someone, an individual, etc)}}\\
         &\cellcolor{gray!10}\multirow{1}{*}{\parbox{\tableColLen}{Questions about a location}}\\
                  & \\
         &  \\
         &\multirow{-3}{*}{\parbox{\tableColLen}{Questions about something numerical (weight, price, any other number)}}\\

    \end{tabular}
    \caption{Prompt patterns for the zero-shot strategies for ChatGPT and Llama. The prompt is of the form ``Here are some examples of \$CLASS\_DESC from a dataset of \$DATASET\_DESC: \$EXAMPLES. Can you create 10 more sentences of that class that would fit the dataset ''. }
    \label{tab:PromptsGen}
\end{table}

We referred to the description given in the original papers of each dataset to craft informative prompts.

\section{Supplementary Results}
\label{Appendix:SuppResults}

\addtocounter{table}{1}
\begin{table*}[]
    \centering
% \vspace{em}
    \begin{tabular}{lccccccc} \toprule
 & SST2 & Irony & IronyB & TREC6 & SNIPS & Average \\ \midrule
Baseline& 87.7/88.8& 67.0/68.1& 43.9/45.9& 84.5/87.9& 95.6/96.2& 75.8/77.4\\
Perfect& 88.9*/89.7*& 67.3/71.1*& 46.0*/48.5*& 87.6*/89.3& 96.1*/96.9*& 77.2/79.1\\
Copy& 87.7/88.8& \textbf{66.2}/69.2& 43.7/46.7& 83.8/86.3& 95.5/\textbf{96.4}& 75.4/77.5\\ \midrule
EDA& 88.1/88.9& 65.9/69.4& 43.6/47.2*& 82.5/86.1& 95.8/96.3& 75.2/77.6\\
AEDA& 88.0/89.0& 65.1/69.5& 42.8/46.7& 82.7/87.4& \textbf{96.1}*/96.3& 74.9/\textbf{77.8}\\
BT& 88.0/89.2*& 64.9/68.4& 44.2/45.8& 83.1/87.1& 95.5/96.3& 75.1/77.3\\
CBERT& 87.6/88.6& 64.8/\textbf{69.5}& 43.6/46.1& 82.9/84.7& 95.4/96.3& 74.9/77.0\\
CBART& 87.7/88.7& 66.0/68.8& 43.6/45.1& 81.2/82.5& 95.9/96.2& 74.9/76.3\\
T5& 87.5/88.8& 64.9/67.7& 43.6/45.6& 82.9/84.8& 95.8/96.3& 75.0/76.6\\
GPT3.5-P& \textbf{88.2}*/\textbf{89.3}*& 65.1/\textbf{69.5}*& \textbf{44.3}/\textbf{47.5}& \textbf{84.5}/\textbf{86.8}& 95.5/96.1& \textbf{75.5}/\textbf{77.8}\\
GPT3.5-ZS& 87.7/88.8& 64.2/67.4& 42.0/45.1& 81.9/86.0& 95.5/95.8& 74.3/76.6\\
GPT3.5-3S& 88.1*/88.6& 64.9/69.0& 40.4/44.4& 83.0/86.8& 95.9/96.2& 74.5/77.0\\
Llama2-P& 87.8/88.9& 65.4/68.1& 44.1/46.1& 83.3/85.9& 95.7/\textbf{96.4}& 75.3/77.1\\
Llama2-ZS& 88.1*/88.8& 65.0/69.2& 42.5/45.7& 82.2/86.0& 95.6/96.0& 74.7/77.2\\
Llama2-3S& 87.8/88.4& 65.0/67.9& 42.8/46.2& 83.8/85.3& 95.7/\textbf{96.4}*& 75.0/76.8\\
\bottomrule
    \end{tabular}
    \caption{Average metric over 15 runs for the training set sizes of 500  (left) and 1000 (right) with a ratio of 1. We report accuracy for binary tasks and macro-f1 for multiclass ones. STDs are between 0.6 and 3.0, depending on the dataset. Stars represent results for which the difference with the baseline was found to be statistically significant.}
    \label{tab:resultsMid}
\end{table*}

\setlength{\tabcolsep}{1pt}

\begin{figure}[]
    \centering
    \tiny
    \begin{subtable}{.48\textwidth}
        \centering
    \begin{tabular}{l*{16}{p{0.35cm}}}
         & \rotatebox{85}{Baseline} & \rotatebox{85}{Perfect}& \rotatebox{85}{Copy} & \rotatebox{85}{EDA} & \rotatebox{85}{AEDA}& \rotatebox{85}{BT} & \rotatebox{85}{CBERT}& \rotatebox{85}{CBART} & \rotatebox{85}{T5} & \rotatebox{85}{GPT3.5-P} & \rotatebox{85}{GPT3.5-ZS} & \rotatebox{85}{GPT3.5-3S} & \rotatebox{85}{Llama2-P} & \rotatebox{85}{Llama2-ZS} & \rotatebox{85}{Llama2-3S}\\
Baseline & \heatmap{0} & \heatmap{0} & \heatmap{0} & \heatmap{20} & \heatmap{10} & \heatmap{10} & \heatmap{30} & \heatmap{20} & \heatmap{20} & \heatmap{0} & \heatmap{40} & \heatmap{30} & \heatmap{20} & \heatmap{30} & \heatmap{30} \\
Perfect & \heatmap{80} & \heatmap{0} & \heatmap{90} & \heatmap{70} & \heatmap{70} & \heatmap{90} & \heatmap{100} & \heatmap{80} & \heatmap{90} & \heatmap{80} & \heatmap{100} & \heatmap{90} & \heatmap{90} & \heatmap{100} & \heatmap{90} \\
Copy & \heatmap{0} & \heatmap{0} & \heatmap{0} & \heatmap{0} & \heatmap{0} & \heatmap{0} & \heatmap{0} & \heatmap{30} & \heatmap{10} & \heatmap{0} & \heatmap{40} & \heatmap{20} & \heatmap{0} & \heatmap{10} & \heatmap{20} \\
EDA & \heatmap{10} & \heatmap{0} & \heatmap{0} & \heatmap{0} & \heatmap{0} & \heatmap{0} & \heatmap{0} & \heatmap{20} & \heatmap{20} & \heatmap{0} & \heatmap{50} & \heatmap{20} & \heatmap{10} & \heatmap{10} & \heatmap{10} \\
AEDA & \heatmap{10} & \heatmap{0} & \heatmap{0} & \heatmap{10} & \heatmap{0} & \heatmap{10} & \heatmap{20} & \heatmap{20} & \heatmap{20} & \heatmap{20} & \heatmap{40} & \heatmap{30} & \heatmap{20} & \heatmap{20} & \heatmap{30} \\
BT & \heatmap{10} & \heatmap{0} & \heatmap{0} & \heatmap{0} & \heatmap{0} & \heatmap{0} & \heatmap{20} & \heatmap{20} & \heatmap{10} & \heatmap{0} & \heatmap{20} & \heatmap{20} & \heatmap{0} & \heatmap{20} & \heatmap{10} \\
CBERT & \heatmap{0} & \heatmap{0} & \heatmap{0} & \heatmap{0} & \heatmap{0} & \heatmap{0} & \heatmap{0} & \heatmap{10} & \heatmap{10} & \heatmap{0} & \heatmap{30} & \heatmap{20} & \heatmap{10} & \heatmap{0} & \heatmap{10} \\
CBART & \heatmap{0} & \heatmap{0} & \heatmap{0} & \heatmap{0} & \heatmap{0} & \heatmap{0} & \heatmap{10} & \heatmap{0} & \heatmap{0} & \heatmap{0} & \heatmap{20} & \heatmap{10} & \heatmap{0} & \heatmap{0} & \heatmap{0} \\
T5 & \heatmap{0} & \heatmap{0} & \heatmap{0} & \heatmap{0} & \heatmap{0} & \heatmap{0} & \heatmap{0} & \heatmap{10} & \heatmap{0} & \heatmap{0} & \heatmap{20} & \heatmap{10} & \heatmap{0} & \heatmap{10} & \heatmap{0} \\
GPT3.5-P & \heatmap{30} & \heatmap{0} & \heatmap{20} & \heatmap{20} & \heatmap{10} & \heatmap{10} & \heatmap{30} & \heatmap{40} & \heatmap{50} & \heatmap{0} & \heatmap{60} & \heatmap{30} & \heatmap{0} & \heatmap{40} & \heatmap{20} \\
GPT3.5-ZS & \heatmap{0} & \heatmap{0} & \heatmap{0} & \heatmap{0} & \heatmap{0} & \heatmap{0} & \heatmap{0} & \heatmap{10} & \heatmap{0} & \heatmap{0} & \heatmap{0} & \heatmap{10} & \heatmap{0} & \heatmap{0} & \heatmap{0} \\
GPT3.5-3S & \heatmap{10} & \heatmap{0} & \heatmap{10} & \heatmap{0} & \heatmap{0} & \heatmap{0} & \heatmap{10} & \heatmap{10} & \heatmap{10} & \heatmap{10} & \heatmap{30} & \heatmap{0} & \heatmap{0} & \heatmap{0} & \heatmap{0} \\
Llama2-P & \heatmap{0} & \heatmap{0} & \heatmap{0} & \heatmap{0} & \heatmap{0} & \heatmap{0} & \heatmap{10} & \heatmap{10} & \heatmap{0} & \heatmap{0} & \heatmap{30} & \heatmap{20} & \heatmap{0} & \heatmap{10} & \heatmap{10} \\
Llama2-ZS & \heatmap{10} & \heatmap{0} & \heatmap{10} & \heatmap{0} & \heatmap{0} & \heatmap{0} & \heatmap{0} & \heatmap{10} & \heatmap{10} & \heatmap{0} & \heatmap{20} & \heatmap{10} & \heatmap{10} & \heatmap{0} & \heatmap{20} \\
Llama2-3S & \heatmap{10} & \heatmap{0} & \heatmap{0} & \heatmap{0} & \heatmap{0} & \heatmap{0} & \heatmap{10} & \heatmap{10} & \heatmap{0} & \heatmap{10} & \heatmap{20} & \heatmap{20} & \heatmap{0} & \heatmap{10} & \heatmap{0} \\

    \end{tabular}
    % \caption{Results of the statistical tests for the starting sizes of 500/1000.}
    \end{subtable}
    \caption{Percentage of times the row algorithm performs statistically better than the column algorithm, with a p-value threshold of 0.05 and using a two-tails paired t-test and with the starting sizes of 500/1000.}
    \label{tab:fewShotStats500}
\end{figure}
% \section{Supplementary Results}
% \label{Appendix:SuppResults}
% In this section we give supplementary results to the paper. Table~\ref{tab:datasets} gives some information about the datasets, and Table~\ref{tab:resultsFull}, the results of data augmentation on the full training set, with a ratio of generated-to-genuine of one.

% Figure~\ref{fig:ratio} shows the performance as we increase the ratio for the GPT3.5-ZS strategy, compared to AEDA and T5, and with a dataset size of 10. As we can observe, the performance plateau quickly for all algorithms. Given that Chat-GPTDesc performs much better on SST-2 than the other datasets, we also give in Figure~\ref{fig:ratioNoSST2} the results while excluding SST-2. We leave for future work to investigate whether the plateauing for GPT3.5-ZS is due to lack of fine-tuning or simply the limit of ChatGPT when it comes to generate diverse sentences.

\end{document}